\documentclass[runningheads]{llncs}
\usepackage{graphicx}
\usepackage{float}
\usepackage{subfig}
\usepackage{multirow}
\usepackage{booktabs}
\begin{document}
\title{Enhanced Optic Disk and Cup Segmentation with Glaucoma Screening from Fundus Images using Position encoded CNNs }
\titlerunning{Optic Disk and Cup Segmentation with Glaucoma Screening}
\author{Vismay Agrawal*\orcidID{0000-0002-4092-3374} \and
Avinash Kori*\orcidID{0000-0002-5878-3584} \and
Varghese Alex*\orcidID{0000-0001-5095-2358}\and
Ganapathy Krishnamurthi*\orcidID{0000-0002-9262-7569} \thanks{All authors contributed equally.}}
\authorrunning{Vismay et al.}
\institute{Department of Engineering Design, Indian Institute of Technology Madras, India\\
\email{gankrish@iitm.ac.in}}
\maketitle              % typeset the header of the contribution
\begin{abstract}
In this manuscript, we present a robust method for glaucoma screening from fundus images using an ensemble of convolutional neural networks (CNNs). The pipeline comprises of first segmenting the optic disk and optic cup from the fundus image, then extracting a patch centered around the optic disk and subsequently feeding to the classification network to differentiate the image as diseased or healthy. In the segmentation network, apart from the image, we make use of spatial co-ordinate (X \& Y) space so as to learn the structure of interest better. The classification network is composed of a DenseNet201 and a ResNet18 which were pre-trained on a large cohort of natural images. On the REFUGE validation data (n=400), the segmentation network achieved a dice score of 0.88 and 0.64 for optic disc and optic cup respectively. For the tasking differentiating images affected with glaucoma from healthy images, the area under the ROC curve was observed to be 0.85. 
\end{abstract}
\section{Introduction}
Glaucoma leads heightened intra-ocular pressure which in-turn leads to damaging of optic nerve head in the eye. Glaucoma during the early stages manifests little or no symptoms, however as the disease progresses symptoms vary from loss of peripheral vision to total blindness. Early detection and treatment of glaucoma helps in preventing the progression of the disease. 
\par In a clinical setup, digital fundus images and optical coherence tomography (OCT) are the often used technique to detect the presence of glaucoma. The optic cup to disk ratio (CDR) is the measure that is commonly computed so as to classify a subject as healthy or diseased. Thus segmenting the optic disk and optic cup is the preliminary step for appropriate classification. 
\par For a variety of classification and segmentation related task, convolutional neural networks (CNNs) have yielded state of the art of performance. In this manuscript, we make use CNN for both the segmentation and the classification task. For the segmentation of optic disk and optic cup from the input, apart from the RGB image, the network is fed the coordinate space (X \& Y) as an additional input to effectively learn the structure of interest. An ensemble of networks was used for both the tasks to reduce the variance in the prediction associated with each model.

%\begin{itemize}
 %   \item Neation fed for doing it with some stats
  %  \item Related work
%\item bias variance trade off, address of intensity homoginety we use CLAHE. 
%\item PRE-TRAINED FOR CLASSIFICATION AND DENSELY CONNECTED FULLY CONVOLTUTIONAL NETWORK FOR SEGMENTATION
%\end{itemize}

\section {Material and Methods}
\begin{figure}
    \centering
    \includegraphics[width = 1.0 \textwidth]{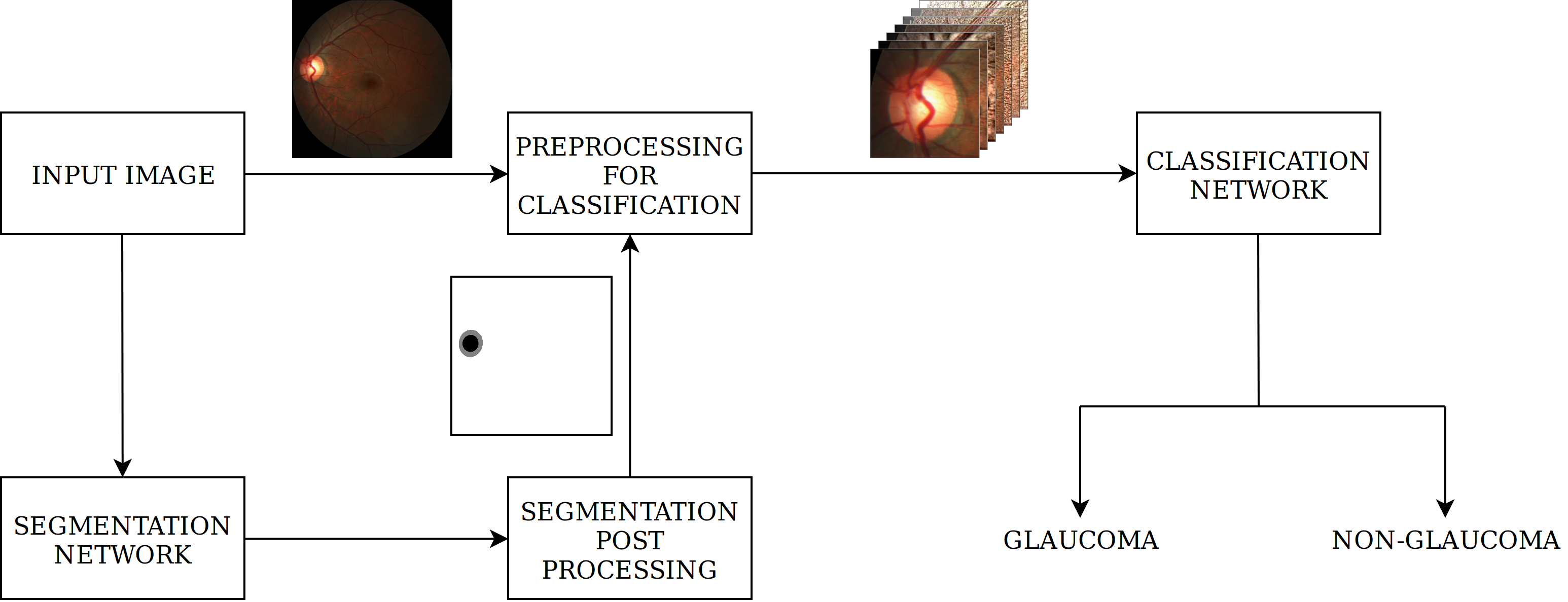}
    \caption{Overall pipeline for segmentation and classification of fundus images using Convolutional neural networks.}
    \label{fig:pipeline}
\end{figure}

The entire pipeline for segmentation of optic disk and cup from fundus images followed by classification is appropriately illustrated in Figure \ref{fig:pipeline}.

\subsection{Data}
In our experiments, we used two publicly available retinal fundus image datasets for glaucoma screening. The REFUGE dataset \cite{refuge}, was made available as part of the Retinal Fundus Glaucoma Challenge (REFUGE). The training data comprises of  400 images centered at the posterior pole with both macula and optic disc, taken using Zeiss Visucam 500 (2124x2056 pixels). The number of glaucoma patients in the dataset is given in Table \ref{first_table}. To further increase the number of data-points, we make use of a publicly available dataset viz DRISHTI-GS1 \cite{DS}. This dataset consists of  101 images which were acquired with the FOV of 30 degrees centered on the OD.

Table \ref{first_table} tabulates the number of glaucoma and healthy fundus images in each dataset. The final dataset comprised of 110 and 391 images under the class glaucoma and healthy respectively. 
\begin{table}[]
\centering
\caption{Distribution of classes namely glaucoma and healthy in the training, validation and testing set across various databases. }\begin{tabular}{@{}cccc@{}}
\toprule
Set        & REFUGE & DRISHTI-GS1 & Total   \\ \midrule
Training    & 28/288 & 70/31   & 98/319  \\
Validation & 6/36   & -       & 6/36   \\
Test     & 6/36   & -       & 6/36   \\ \midrule
Total          & 40/360 & 70/31   & 110/391 \\ \bottomrule
\end{tabular}

\label{first_table}
\end{table}

\subsection{Convolutional Neural Network}
Convolution Neural Networks(CNNs) was first introduced by Fukushima. When compared to the network first introduced, the current CNN architectures are deeper and comprises of more number of convolutions, skip \& residual connections.  

\subsection{Segmentation  network}
\begin{figure}
    \centering
    \includegraphics[width = 0.6\textwidth]{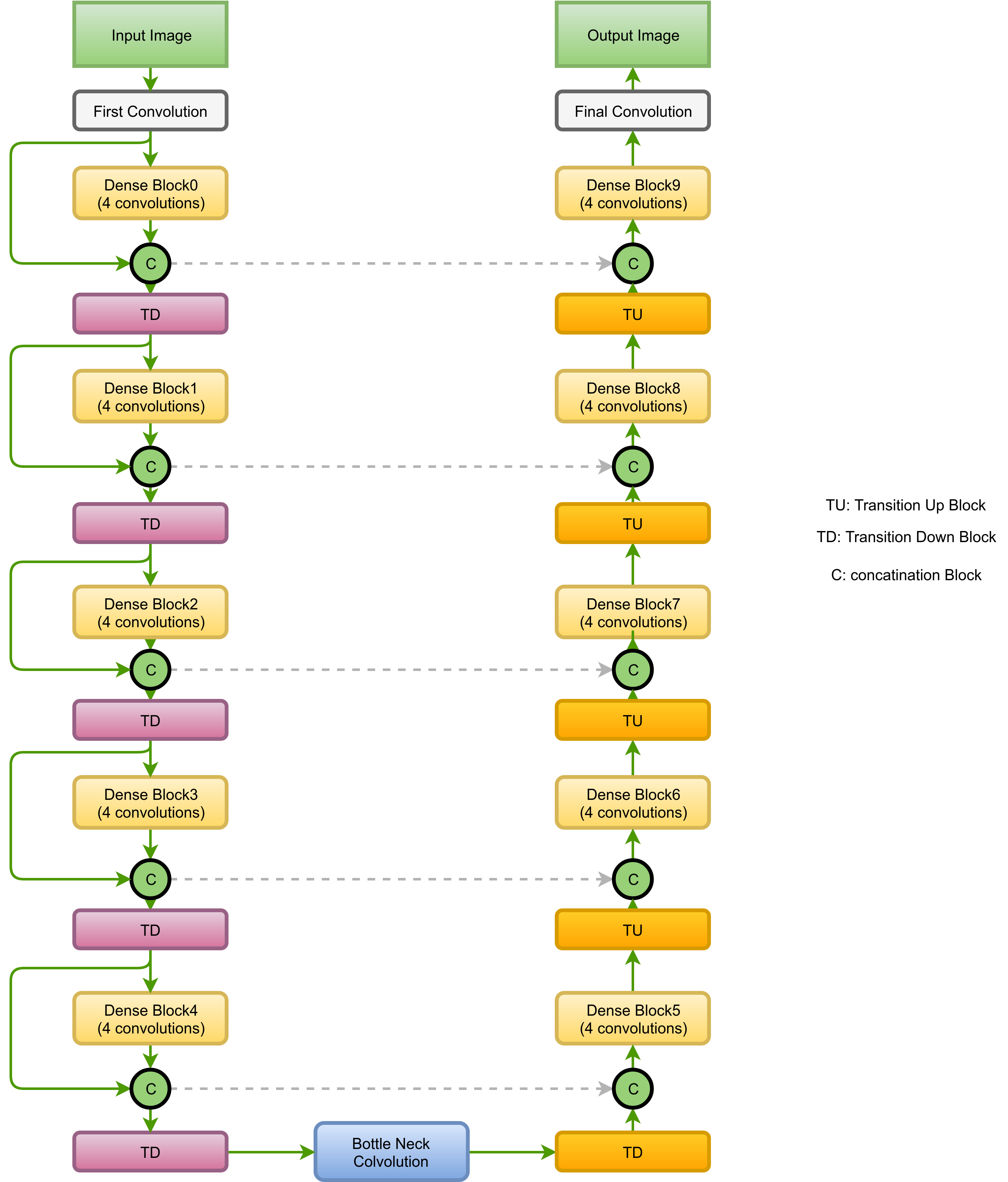}
    \caption{Architecture of the Segmentation Network.}
    \label{segnet}
\end{figure}

For the segmentation task, a 57 layered deep densely connected semantic segmentation convolutional neural network was used, Fig. \ref{segnet}. The network comprises of long and short skip connections so as to effective reuse the features learned by the network. 

\subsubsection{Pre-Processing of Data}
Images were cropped to nearest square size and resized to a dimension of (512, 512).  The different lighting conditions and intensity variations among images across various databases were circumvented by performing normalization of the histogram using Contrast Limited Adaptive Histogram Equalization (CLAHE).  Two different images were generated by varying parameters such as clip value \& window level while performing CLAHE. The value of hyper-parameters associated with CLAHE used in this manuscript is given in Table \ref{segClahe}. Along with CLAHE, spatial co-ordinates information were also provided to the network. This additional information aided in learning relative features (i.e. disk location with respect to fovea). 
 
%One of the networks was fed with the original, 2 CLAHE images and spatial co-ordinates  (total 11 channels) as input, and other with just the original, and spatial co-ordinate matrix (total 5 channels).

\begin{table}[]
\centering
\caption{clahe parameters for segementation}
\begin{tabular}{@{}ccc@{}}
\toprule
\multicolumn{1}{l}{S.No.} & \begin{tabular}[c]{@{}c@{}}Tile Grid Size\\ (w, h)\end{tabular} & Clip Limit \\ \midrule
1                         & (8,8)               & 2          \\
2                         & (300, 300)          & 2          \\ \bottomrule
\end{tabular}
\label{segClahe}
\end{table}

\subsubsection{Ensemble of segmentation networks}
The ensemble was generated by clubbing 2 different segmentation network, which differs in the number of channels provided as the input. 
One of the networks was trained with the original image and spatial co-ordinate matrix (total 5 channels) as the inputs while the other was trained by appending 2 CLAHE images in the same set of inputs (total 11 channels).

\subsubsection{Training}
The dataset was split into training, validation and testing in the ratio 70:20:10. Both the networks were trained and validated on 278 and 59 images respectively. To further address the issue of class imbalance in the network, the parameters of the network were trained by minimizing weighted cross entropy. The weight associated to each class was equivalent to the ratio of the median of the class frequency to the frequency of the class of interest. To increase the number of data points on the fly data augmentation was used with random rotation between 0-360 degree, and random flip with a probability of 0.5 was also implemented. The number of batches was set at 4, while the learning rate was initialized to 0.0001 and decayed by a factor of 10 \% every-time the validation loss plateaued. The network was trained for 50 epochs using weighted cross entropy as a cost function, weights were estimated by using entire training data as a ratio of the median of frequency

\subsubsection{Inference}
The normalized data was fed to both the network \& each network predicts the returns a segmentation mask. The final segmentation was based on computing the arithmetic mean of segmentation generated by the networks.

\subsubsection{Post Processing}
\begin{figure}
    \centering
    \includegraphics[width = 1.0 \textwidth]{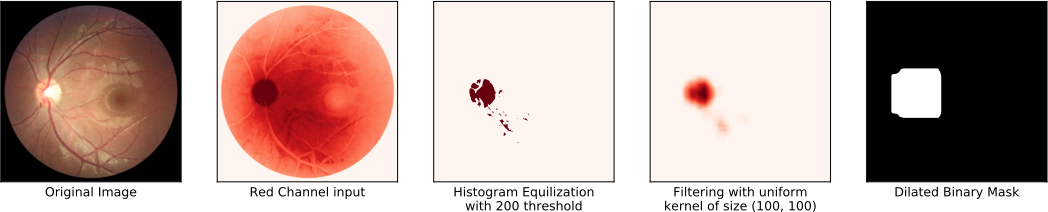}
    \caption{Steps involved in generating masks}
    \label{threshseg}
\end{figure}

Rough masks of the optic disk were estimated using traditional computer vision techniques with the help of image filtering and thresholding, Figure \ref{threshseg}. The segmentation's attained by the network were then multiplied with these masks to remove false positives. Maximum area ellipse was fit on individual class (optic disk and cup) and fused to obtain final prediction. 

\subsection{Classification network}
For the classification task, a DenseNet201 \cite{huang2017densely} \& ResNet18 \cite{he2016identity} pre-trained on natural images \cite{deng2009imagenet} forms the ensemble. The hindmost layer in the network i.e. the classification layer was modified to have 2 neurons. 
\par An additional convolutional layer was appended before both the pre-trained models to convert out 21 channel input to 3 channels. To make the network images accept inputs of variable dimension, the global average pooling layer was substituted with an adaptive average pooling layer.

\subsubsection{Pre-Processing of Data} 
\begin{figure}
    \subfloat[]{\includegraphics[width = 0.14 \textwidth]{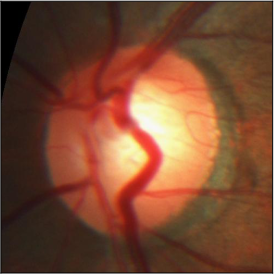}}\hfill
    \subfloat[]{\includegraphics[width = 0.14 \textwidth]{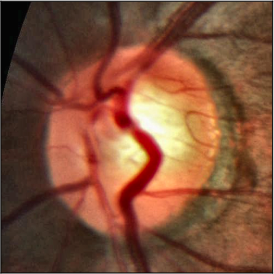}}\hfill
    \subfloat[]{\includegraphics[width = 0.14 \textwidth]{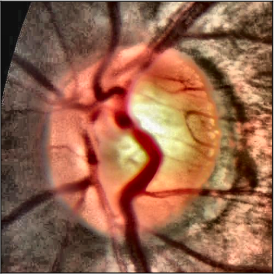}}\hfill
    \subfloat[]{\includegraphics[width = 0.14 \textwidth]{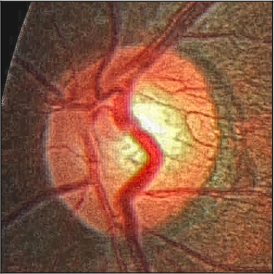}}\hfill
    \subfloat[]{\includegraphics[width = 0.14 \textwidth]{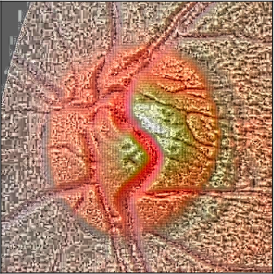}}\hfill
    \subfloat[]{\includegraphics[width = 0.14 \textwidth]{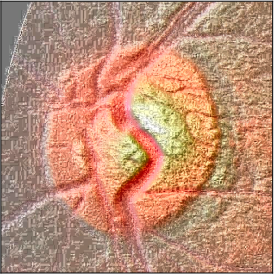}}\hfill
    \subfloat[]{\includegraphics[width = 0.14 \textwidth]{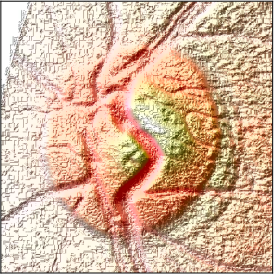}}\\
    \subfloat[]{\includegraphics[width = 0.14 \textwidth]{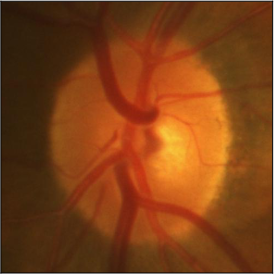}}\hfill
    \subfloat[]{\includegraphics[width = 0.14 \textwidth]{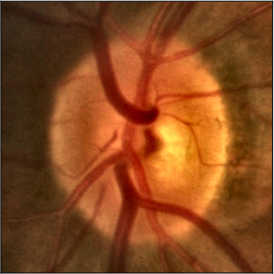}}\hfill
    \subfloat[]{\includegraphics[width = 0.14 \textwidth]{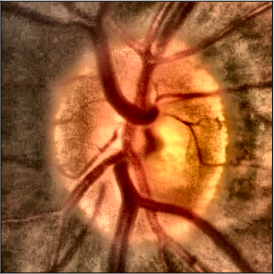}}\hfill
    \subfloat[]{\includegraphics[width = 0.14 \textwidth]{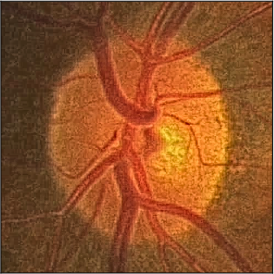}}\hfill
    \subfloat[]{\includegraphics[width = 0.14 \textwidth]{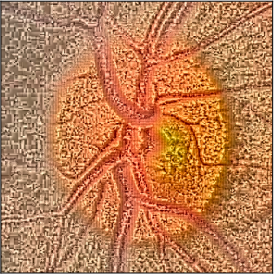}}\hfill
    \subfloat[]{\includegraphics[width = 0.14 \textwidth]{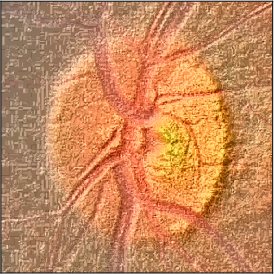}}\hfill
    \subfloat[]{\includegraphics[width = 0.14 \textwidth]{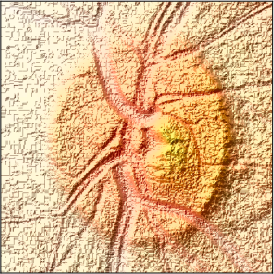}}\\
    \caption{Row 1 and 2 consists of glaucoma and non glaucoma optic disk images respectively with corresponding CLAHE images as the parameter is varied according to the Table \ref{classificationClahe}}
    \label{fig:clashe}
\end{figure}

The pixel level segmentation of the optic disk and optic cup was utilized to generate images of dimension 550 $\times$ 550 centered around the optic disk. 6 different images were generated by varying parameters such as clip value \& window level while performing CLAHE. The various parameters used for CLAHE are illustrated in Table \ref{classificationClahe}. The network is fed the original and 6 histogram equalized images (i.e. 21 channels) as input. Figure \ref{fig:clashe} (a-g) \& (h-n) illustrates the variation in a glaucoma and healthy/normal image upon varying the hyper-parameters associated to CLAHE.

\subsubsection{Training}
Stratified sampling was utilized so as to ensure glaucoma data was present in the training, validation and testing data. The acute class imbalance among the classes was addressed by training the network to optimize weighted cross entropy. The weights begin reciprocal of the frequency of occurrence of each class in the held out training data. The network was trained with a batch size set to 4 with ADAM as the optimizer. The learning rate was initialized to  0.0001  was annealed by a factor of 0.1 at the end of every 7 epochs. The network was trained for a total of 80 epochs. 

\subsubsection{Inference}
The segmentation generated by the trained segmentation network was used to localize the optic disk and extract patch of dimension 550 $\times$ 550. Ten crops of the images of dimension 500 $\times$ 500 were extracted from the patch and fed as input to each model in the ensemble. For a given crop, the model in the ensemble yields the probability of it being glaucoma. The risk of a patient being affected with glaucoma was equivalent to the average of the posterior probabilities.

\begin{table}[]
\centering
\caption{clahe parameters for classification}
\begin{tabular}{@{}ccc@{}}
\toprule
\multicolumn{1}{l}{S.No.} & \begin{tabular}[c]{@{}c@{}}Tile Grid Size\\ (w, h)\end{tabular} & Clip Limit \\ \midrule
1                         & (8,8)                                                           & 2          \\
2                         & (8,8)                                                           & 10         \\
3                         & (100,100)                                                       & 2          \\
4                         & (100,100)                                                       & 100        \\
5                         & (300,300)                                                       & 2          \\
6                         & (500,500)                                                       & 2          \\ \bottomrule
\end{tabular}
\label{classificationClahe}
\end{table}

\begin{figure}[H]
    \subfloat[]{\includegraphics[height = 0.25 \textwidth]{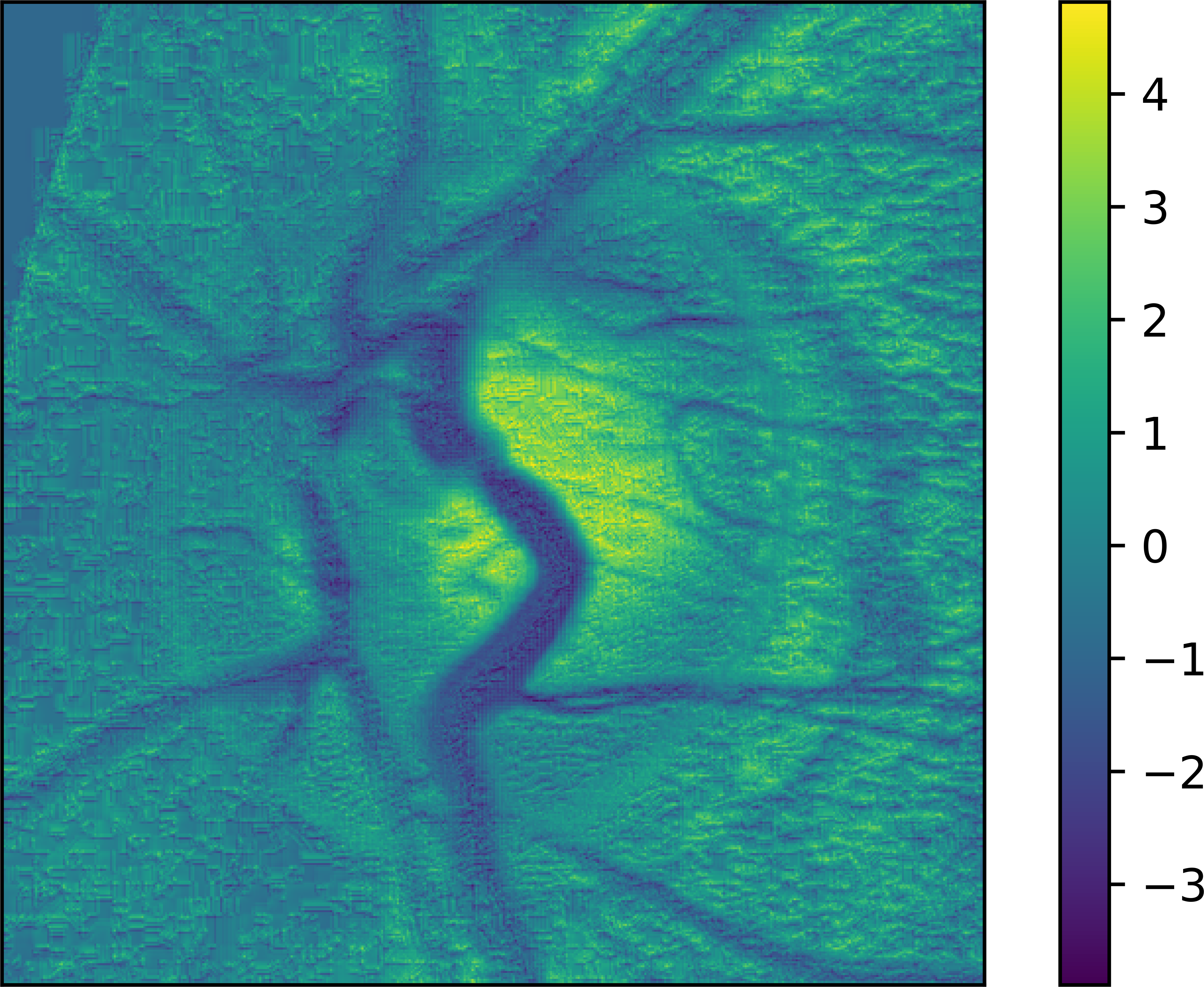}}\hfill
    \subfloat[]{\includegraphics[height = 0.25 \textwidth]{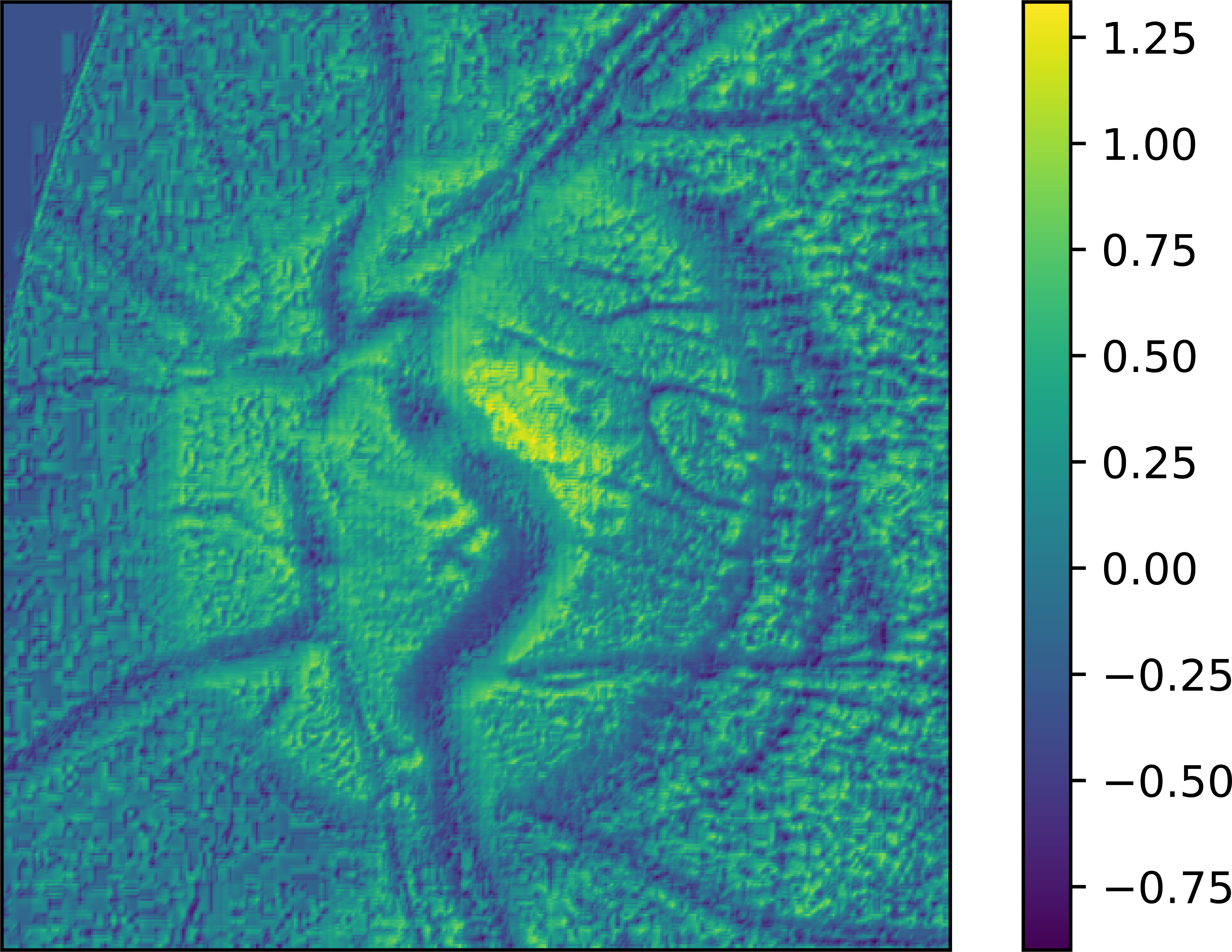}}\hfill
    \subfloat[]{\includegraphics[height = 0.25 \textwidth]{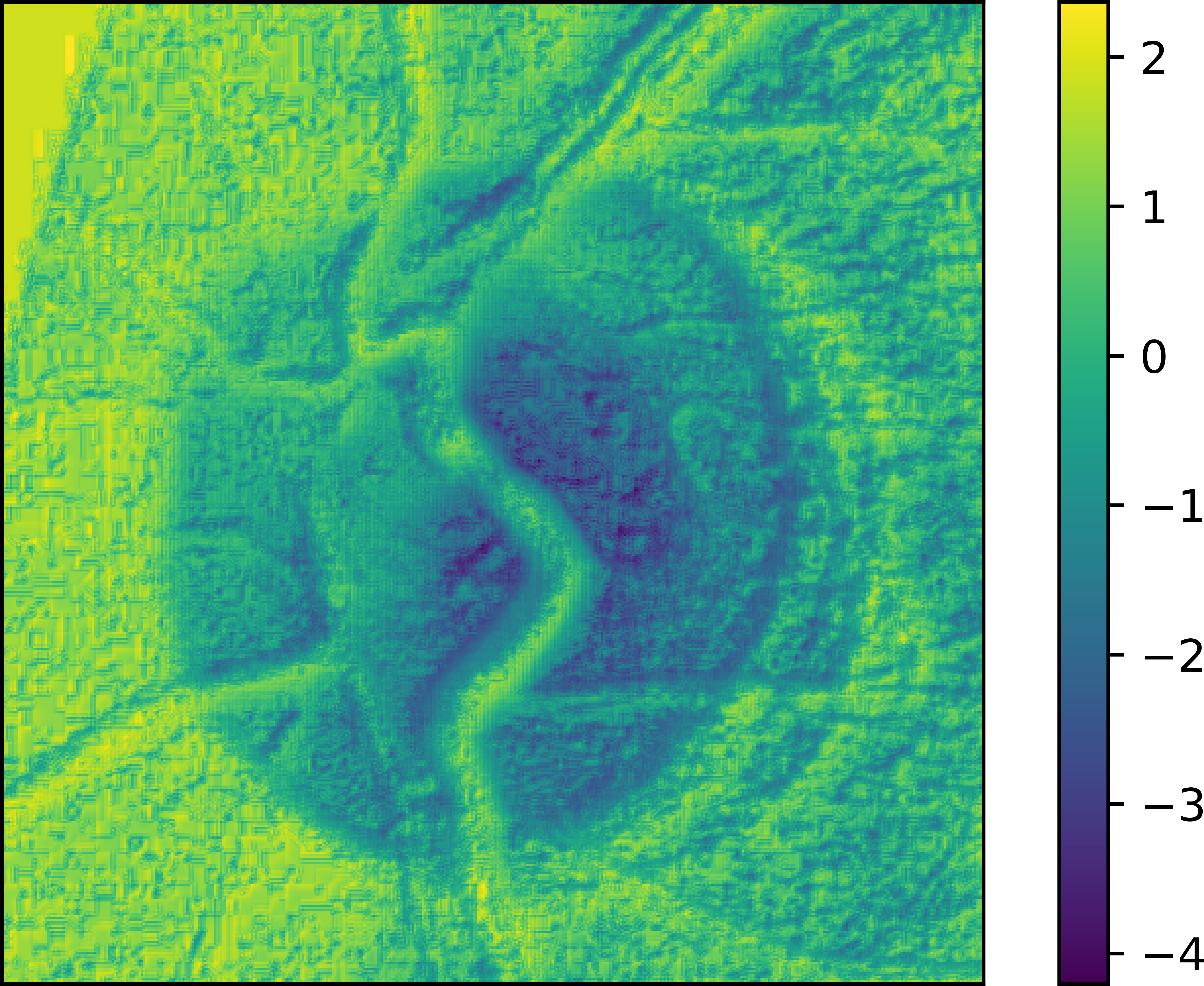}}\\
    \subfloat[]{\includegraphics[height = 0.25 \textwidth]{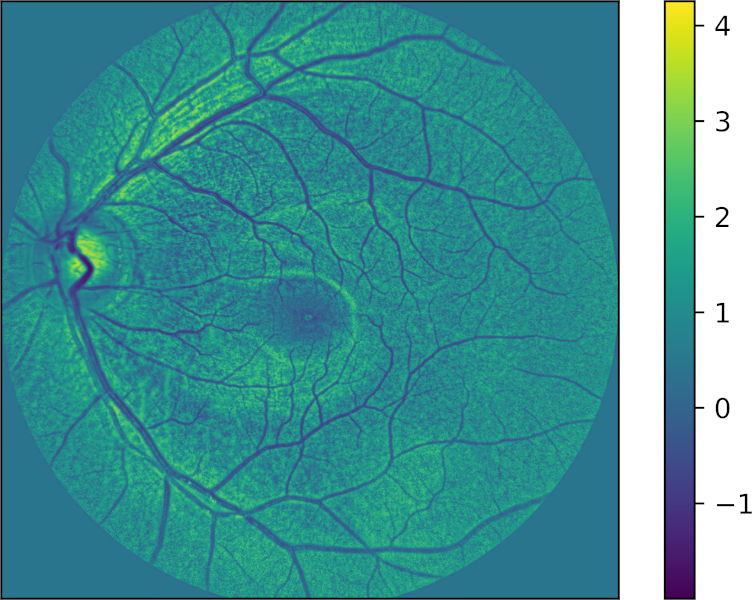}}\hfill
    \subfloat[]{\includegraphics[height = 0.25 \textwidth]{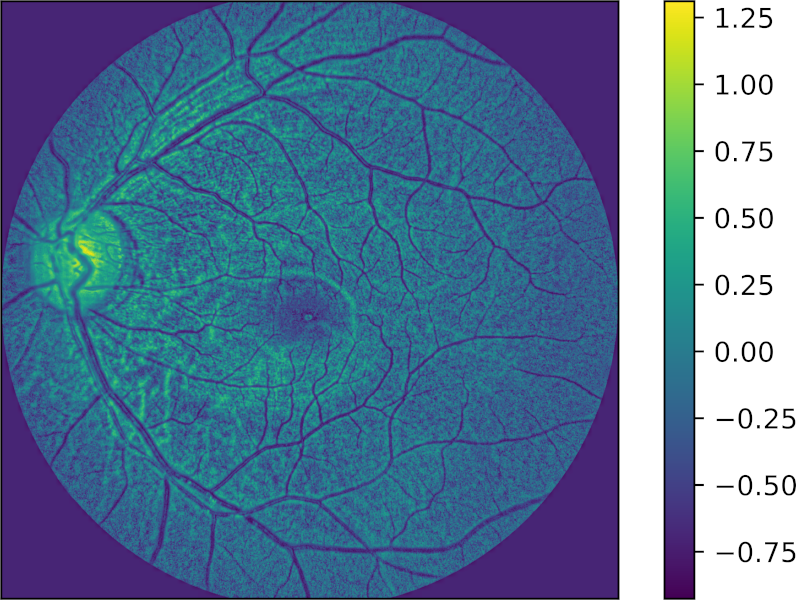}}\hfill
    \subfloat[]{\includegraphics[height = 0.25 \textwidth]{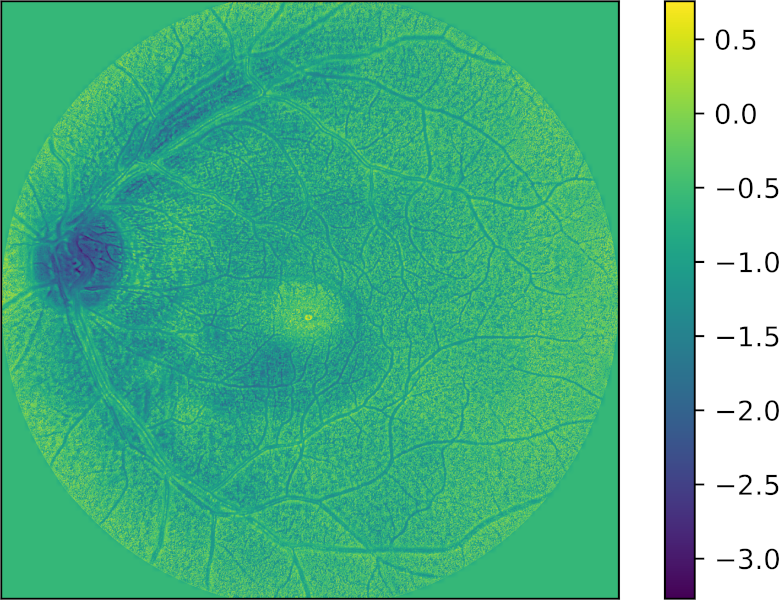}}\\
    \caption{Activation Map generated by the  trained classification network on passing patch centered around optic disk \& entire fundus image. The first feature map (a, d) was observed to assign higher weight-age to the optic cup while the second (b, e) and the third (c, f) feature was observed to assign higher activation upon encountering optic disk \& background respectively in the input image. }
    \label{fig:clahe}
\end{figure}

%The network encoded the qualities of optic disk images by learning different features from each CLAHE image and combining the results. We can infer that our deep neural network is looking for optic disk and optic cup for classification tasks. We can 

\section{Results}
\subsection{Performance of Segmentation network }
The performance of the trained network was gauged by testing on the offline validation data provided by the challenge organizers. The offline data consist of 400 fundus images acquired using a Canon CR-2 with a dimension of 1634 $\times$1634. The proposed segmentation network achieved a dice score of 0.64 and 0.88 for optic cup \& optic disk respectively with a mean absolute error CDR 0.09. 
.

\subsection{Performance of Classification network }

\par On the validation data, the segmentation generated by the segmentation network was used to extract patches centered around OD and fed as input to the trained classification network. The proposed classification network achieved a sensitivity of 0.75 at a specificity of 0.85 and 0.856 area under the ROC curve.   
% \begin{figure}
%     \centering
%     \includegraphics[width = 1.0 \textwidth]{imgs/ROC_CURVE.png}
%     \caption{Above figure shows the classification ROC on held-out test data}
%     \label{AUROC}
% \end{figure}

\subsubsection{Effect of using CLAHE}
On visualizing the output/activation map of the first convolutional layer of the trained classification network, we observe that the network learns to focus on optic cup, optic disk and background from the input. Figure \ref{fig:clahe} (a-c) illustrates the activation map upon feeding a cropped patch as the input. Similar observations were found when the entire fundus was fed in the trained network.

\section{Conclusion}

In this manuscript, we make use of an ensemble of CNNs for first segmenting the optic disc and optic cup from digital fundus images. From the segmentation, patches centered around the optic disc were fed to ensemble of classifiers. Models viz DenseNet-201 and ResNet-18 pre-trained on natural images form the ensemble. The histogram of the fundus was equalized using CLAHE. The major observations we observed were:
\begin{itemize}
\item Ensemble of classifiers/models produces a better performance when compared to using a single model
\item Selective pruning of the ensemble aids in achieving higher accuracy/dice score when compared to using all the models in the ensemble
\item Using multiple CLAHE parameters generalizes better across various datasets, than just using single CLAHE
\item Using position channels in input layer helps in the better localization of optic disk and cup

\end{itemize}
On the offline validation data, the scheme explored in this manuscript generalizes well on unseen data and achieves competitive results. 

%For the task of segmentation of Optic disk and cup :
%\begin{itemize}
%\item Deep network with densely connected layers with skip and residual connection helps in learning better features with less number of parameters
%\item Using multiple CLAHE parameters generalizes better across various datasets, than just using single CLAHE

%\end{itemize}

\bibliographystyle{plain}
\bibliography{ref}

\begin{thebibliography}{1}

\bibitem{refuge}
Refuge.
\newblock \url{http://refuge.grand-challenge.org}.
\newblock Accessed: 2018-07-28.

\bibitem{deng2009imagenet}
Jia Deng, Wei Dong, Richard Socher, Li-Jia Li, Kai Li, and Li~Fei-Fei.
\newblock Imagenet: A large-scale hierarchical image database.
\newblock In {\em Computer Vision and Pattern Recognition, 2009. CVPR 2009.
  IEEE Conference on}, pages 248--255. IEEE, 2009.

\bibitem{he2016identity}
Kaiming He, Xiangyu Zhang, Shaoqing Ren, and Jian Sun.
\newblock Identity mappings in deep residual networks.
\newblock In {\em European Conference on Computer Vision}, pages 630--645.
  Springer, 2016.

\bibitem{huang2017densely}
Gao Huang, Zhuang Liu, Kilian~Q Weinberger, and Laurens van~der Maaten.
\newblock Densely connected convolutional networks.
\newblock In {\em Proceedings of the IEEE conference on computer vision and
  pattern recognition}, volume~1, page~3, 2017.

\bibitem{DS}
Jayanthi Sivaswamy, S~Krishnadas, Arunava Chakravarty, GD~Joshi, A~Syed Tabish,
  et~al.
\newblock A comprehensive retinal image dataset for the assessment of glaucoma
  from the optic nerve head analysis.
\newblock {\em JSM Biomedical Imaging Data Papers}, 2(1):1004, 2015.

\end{thebibliography}
\end{document}